\documentclass[10pt,twocolumn,letterpaper]{article}

\usepackage[pagenumbers]{cvpr} 

\definecolor{cvprblue}{rgb}{0.21,0.49,0.74}
\usepackage[pagebackref,breaklinks,colorlinks,allcolors=cvprblue]{hyperref}

\def\paperID{34} 
\def\confName{CVPR}
\def\confYear{2026}

\title{The Right Inference Strategy Is All You Need: Nearly Training-Free Domain-Wise Inference for EgoCross Challenge\\ \large Team Name: WFJ-KnowinEnvision}

\author{Leyi Wu\textsuperscript{\rm 1,3,$*$}, 
Yifan Zhao\textsuperscript{\rm 1,$*$}, 
Jinjie Zhang\textsuperscript{\rm 1, $*$}, 
Yinchuan Li\textsuperscript{\rm 3}, 
Yingcong Chen\textsuperscript{\rm 1,2,$\dag$},\\
\textsuperscript{\rm 1}HKUST(GZ), \textsuperscript{\rm 2}HKUST, \textsuperscript{\rm 3}Knowin\\
{\tt\small \{lwu398, yzhao642, jzhang103\}@connect.hkust-gz.edu.cn; yingcongchen@ust.hk}
}

\begin{document}
\maketitle

\renewcommand{\thefootnote}{}
\footnotetext{$*$ Equal contribution. $\dag$ Corresponding Author.}
\footnotetext{Technical Report of the \href{https://egocross-benchmark.github.io/}{2026 EgoCross Challenge}}
\begin{abstract}
EgoCross evaluates multimodal large language models on egocentric video question answering under substantial domain shift, where test videos come from surgery, industrial assembly, extreme sports, and animal-mounted cameras rather than ordinary daily-life scenes.
In the source-limited track, the base model is fixed to Qwen3-VL-4B, while the official task-specific support set contains only 20 training samples.
This setting makes the challenge less about model scaling and more about exposing the right visual, temporal, and answer-selection cues to a constrained model.
Our key observation is that the frozen baseline model is not simply incapable of these rare scenarios; rather, it often fails to transfer its existing visual-language knowledge to the new task format without an appropriate interface.
We therefore use a domain-wise inference strategy that treats the four target domains separately and designs different input, prompting, and answer-mapping procedures according to each domain's task characteristics.
These strategies make the rare egocentric scenes more interpretable to the VLM by emphasizing the cues that matter for each domain.
The resulting system is nearly training-free: surgery, and animal questions are answered with the base Qwen3-VL-4B model, while XSports and industry use only the official SFT checkpoint trained for two epochs on the provided 20 training samples.
On the final evaluation, this simple strategy reaches 66.98\% overall accuracy, suggesting that careful domain-aware inference can compensate for limited base-model strength and recover much of the ability already present in the baseline model.\footnote{Code is available at \url{https://github.com/YUEVII/Egocross-Challenge}.}
\end{abstract}    
\section{Introduction}
\label{sec:intro}

Egocentric video question answering requires a model to reason from first-person visual evidence and choose the correct answer from a small set of candidates.
EgoCross~\cite{li2026egocross} makes this setting substantially harder by evaluating multimodal large language models on four novel domains: laparoscopic surgery, industrial assembly, extreme sports, and animal-mounted videos.
These domains differ from common daily-life videos in camera motion, viewpoint, object vocabulary, temporal granularity, and the relation between the camera bearer and the observed action.
The benchmark therefore probes not only generic recognition ability, but also whether a model can transfer its visual-language priors to unfamiliar egocentric contexts.

The source-limited track further changes the nature of the problem because participants cannot simply replace the base model with a larger or more recent system.
All submissions are constrained to Qwen3-VL-4B~\cite{bai2025qwen3} as the base model, and the organizers provide only 20 task-specific training samples.
Under this constraint, the central question becomes: how much of the target-domain performance can be recovered without learning new parameters?
We argue that even a constrained frozen multimodal model can solve many EgoCross~\cite{li2026egocross} questions if the inference interface matches the question's domain and type.
In practice, many failures are caused by avoidable mismatches, such as asking the model to infer a temporal option without explicit frame timing, letting salient background objects compete with the true interaction target, or allowing verbose generation when the evaluator expects a single letter.

We propose a domain-wise inference strategy for EgoCross Challenge~\cite{li2026egocross}.
Instead of applying one universal prompt to all videos, we process the four target domains separately while preserving a shared high-level idea across them.
For each domain, we first identify the dominant source of ambiguity, then choose the visual input format, task cue, and answer-normalization rule that makes the multiple-choice decision easiest for the fixed model.
This design is intentionally lightweight: surgery and animal questions use the base Qwen3-VL-4B model without fine-tuning, while XSports and industry use only the official SFT model trained for two epochs on the provided 20 training samples.
Thus, our system keeps the model backbone essentially unchanged and concentrates adaptation in the inference procedure.

The four domains motivate different concrete inference programs under the same design principle.
For animal-mounted videos, we emphasize interaction targets and deterministic frame-to-time mapping because the low moving camera often makes background objects visually salient.
For surgical videos, we combine direct MCQ generation, visibility verification, coordinate-to-region mapping, and earliest-start detection to match the structure of tool-centric procedural questions.
For industrial assembly, we route different question types to verification, pairwise comparison, counting, and spatial-localization experts.
For XSports, we use specialized temporal/action routers to handle fast motion and closely spaced candidate timestamps.
Together, these domain programs turn rare egocentric scenes into more explicit recognition, grounding, verification, and option-matching problems for the VLM.

Our contribution is a simple but effective recipe for source-limited EgoCross Challenge~\cite{li2026egocross}.
First, we show that a constrained base MLLM can be highly competitive under severe domain shift when the inference interface is carefully aligned with the task.
Second, we use separate domain programs for animal, surgery, industry, and XSports videos, but keep the same design principle of media packaging, domain cueing, and deterministic answer extraction.
Third, our final submission reaches 66.98\% overall accuracy, supporting the view that inference strategy can compensate for limited base-model strength in this setting.
\section{Method}
\label{sec:method}

\subsection{Overview}
\label{subsec:method_overview}

We treat EgoCross~\cite{li2026egocross} as a domain-wise multiple-choice inference problem over a fixed VLM.
Given a sequence of sampled frames, a question, and four answer options, the system must output exactly one letter from A to D.
Our method keeps Qwen3-VL-4B~\cite{bai2025qwen3} as the common base model and adapts only the inference program.
Each domain program specifies three elements: how the visual evidence is packaged for the model, which lightweight cue is added to reduce domain-specific ambiguity, and how the generated text is converted into a valid multiple-choice answer.

The four target domains are handled separately because their dominant ambiguities are different, but the high-level strategy is shared.
Surgical videos emphasize fine-grained tool and action recognition, industrial videos emphasize object-operation grounding, XSports videos emphasize fast motion and temporal anticipation, and animal videos emphasize interaction grounding from a low and unstable camera viewpoint.
For surgery and animal, we use the base model directly without any parameter update.
For XSports and industry, we use the official source-limited SFT checkpoint trained for two epochs on the provided 20 training samples, because these domains benefit from the support examples for fast-motion action transitions and industrial object-operation grounding.
All predictions are produced with deterministic decoding and then normalized to the official answer format.

\subsection{Animal Domain}
\label{subsec:animal_domain}

\paragraph{Common setup.}
For the animal domain, we use the frozen Qwen3-VL-4B~\cite{bai2025qwen3} base model and dispatch each EgoPet~\cite{li2026egocross} question according to its official question type: animal identification, interaction identification, or interaction temporal localization.
All three branches share the same lightweight inference setup.
We use the provided frame sequence in chronological order and apply deterministic decoding to keep the answer format stable.
For answer selection, branches that directly ask for A, B, C, or D use the generated letter as the prediction, while temporal localization first grounds the queried event to a frame and then maps it to the corresponding option interval.

\paragraph{Animal identification.}
Animal identification questions are answered with the model's native video interface.
These questions usually require recognizing the animal, object, or scene-level visual evidence across the clip, so preserving temporal continuity is more useful than attaching additional task-specific instructions to every frame.
We therefore package the ordered frames as a single video input with a sampling rate of 1.0 FPS and pair it with a plain multiple-choice prompt containing the question, the four normalized options, and the instruction to answer only with a single letter.
No domain-specific hint is added in this branch; the goal is to let the base model use its general visual recognition ability while constraining the output space to A--D.

\paragraph{Interaction identification.}
Interaction identification questions are answered with a multi-image prompt because the key evidence is often a local relation between the animal and a nearby target.
Pet-mounted videos contain strong camera motion and many close background objects, so the most salient object in the frame is not necessarily the object being asked about.
To reduce this ambiguity, we keep the official 0.5 FPS frame order explicit and prepend each frame with its chronological time label, where image $i$ is associated with an approximate two-second window.
The prompt then uses a negative interaction cue: the model is asked to choose the object the animal is actively engaging with, and not to choose a large or nearby background object unless the animal touches, chases, eats, sniffs, watches closely, or plays with it.
This branch still directly asks for a single A--D answer, keeping the inference format consistent with the official multiple-choice evaluation.

\paragraph{Interaction temporal localization.}
Interaction temporal localization questions are reformulated from option selection into event grounding followed by deterministic option matching.
Instead of asking the model to directly choose among temporal intervals, we show the official 0.5 FPS frame sequence as ordered images with phase-aware labels: each frame receives both its approximate time window and a coarse clip region label such as earliest, early, middle, late, or latest.
The model is instructed to return the earliest image number where the event in the question first clearly occurs or begins.
Given the predicted image index $i$, we use the midpoint of the corresponding official frame window as the representative timestamp and select the option interval that best matches it.
When the predicted index exceeds the number of available frames, we select the option with the latest ending time.
Because the test frames follow a fixed FPS ($0.5$), the frame-to-time conversion is deterministic; this lets the VLM focus on grounding when the event first appears, instead of additionally performing time conversion and interval-to-option matching.

\subsection{Surgery Domain}
\label{subsec:surgery_domain}

\paragraph{Common setup.}
For the surgical domain, we use the frozen Qwen3-VL-4B base model and apply dataset-specific inference programs for CholecTrack20 and EgoSurgery.
Both programs operate directly on the benchmark-released frame sequences, preserve their chronological order, and use deterministic decoding only.
Rather than relying on one uniform prompt across all surgical questions, we adapt the inference procedure to the structure of each question type.
The resulting design combines direct MCQ generation for question types whose answer is visually explicit, yes/no verification for visibility and dominant-tool judgments, coordinate regression for spatial localization, and a dedicated earliest-start detector for CholecTrack20 temporal localization.

\paragraph{EgoSurgery.}
EgoSurgery is handled entirely with a base-model inference program.
When temporal reasoning requires a timestamp interpretation, we follow the benchmark's released frame spacing for this dataset rather than assuming access to the original video.
Object counting is answered with direct MCQ generation rather than a specialized counting verifier, because in this dataset the direct multiple-choice prompt was empirically more reliable than the option-guided alternative.
Object-not-visible identification is handled by chunked yes/no verification: for each candidate instrument, the model scores whether that instrument is clearly visible anywhere in the segment, and we choose the option with the smallest visibility score.
Dominant held-object identification uses a similar verification style, but aggregates yes/no evidence across frame chunks with a frame-weighted mean margin, selecting the tool that the queried hand interacts with for the largest share of time.
Object spatial localization first restricts the evidence to a small temporal neighborhood around the queried timestamp, then asks the model to output seven permille-coordinate points on the target instrument; the median point is mapped to the nearest answer region.

\paragraph{CholecTrack20: common setup.}
CholecTrack20 uses the same frozen base model, but requires a more strongly typed inference program because the benchmark mixes temporal localization, next-step anticipation, and fine-grained tool reasoning within the same surgical workflow.
We therefore use one inference routine for each question family instead of forcing all questions through a single prompt.
In practice, dominant-tool identification, object-not-visible identification, object counting, and spatial localization are handled with base-model visual reasoning programs, while next-action prediction, next-phase prediction, and action temporal localization each use task-specific temporal inference logic.

\paragraph{CholecTrack20 base router.}
For CholecTrack20, timestamp interpretation follows the frame spacing implied by the released benchmark frames and question wording; in the historical benchmark configuration, VID25 and VID111 are treated with a denser effective spacing than the default setting.
Dominant held-object identification, object-not-visible identification, and object spatial localization follow the same high-level strategies as EgoSurgery: frame-weighted yes/no verification for dominant tool usage, chunked visibility verification for not-visible questions, and coordinate-to-region mapping for spatial localization.
Object counting uses option-guided verification over the four numeric answer choices rather than direct generation.
For the temporal-localization branch used inside the general CholecTrack20 solver, we first restrict the visual evidence to the released frames nearest to the four candidate timestamps and then apply direct MCQ decoding on this time-focused frame set.

\paragraph{CholecTrack20 next-action and next-phase prediction.}
Next-action prediction and next-phase prediction are handled by dedicated direct-MCQ branches specialized to procedural anticipation.
Both branches retain the official answer format, but prepend a surgery-context prompt that reminds the model of the canonical laparoscopic cholecystectomy workflow.
This prompt is especially useful when the visible tail frames are compatible with multiple local motions but only one continuation is plausible under the global phase order.
The next-action branch specializes to immediate procedural action prediction, while the next-phase branch specializes to the coarser workflow transition.

\paragraph{CholecTrack20 action temporal localization.}
For action temporal localization, we use a dedicated ATL2 earliest-start detector instead of relying on direct MCQ decoding alone.
ATL2 first obtains a baseline timestamp-to-option prediction by asking the model for the earliest frame index where the queried action starts and mapping that frame to the nearest option timestamp.
It then refines this prediction with an earliest-positive scan over the four candidate timestamps.
For each candidate timestamp $t$, we gather a small local panel around $t-1$, $t$, and $t+1$ seconds, ask whether the target action has already started at this timestamp, and score the response with yes/no next-token probabilities.
After sorting the candidates by time, the earliest timestamp whose yes probability exceeds a fixed threshold is selected; when no candidate passes the threshold, or when the earliest positive is insufficiently separated from the next one, the branch falls back to the baseline answer.
This design encourages the model to decide whether the action has begun \emph{by} a candidate time, rather than to directly compare four nearby timestamps in one generation step, which was empirically less stable on this dataset.

\subsection{Industry Domain}
\label{subsec:industry_domain}

\paragraph{Common setup.}
For the industry domain, we run a dedicated router over ENIGMA questions and leave all non-ENIGMA submission rows blank in the domain-specific output file.
The router executes nine deterministic component experts and then merges their predictions by question type: two SFT experts for action temporal localization, one base-model expert for dominant held-object identification, one base-model pairwise expert for next interaction prediction, three base-model experts for object counting, one SFT expert for object-not-visible identification, and one base-model expert for object spatial localization.
All components resolve the question-specific frame paths under the test set, use an effective sampling rate of 0.5 FPS, attach chronological frame-time prefixes, and cap each call at at most 10 effective frames unless a strategy first restricts the frames around a queried timestamp.
Direct generation is greedy, while the verification-based branches choose answers from next-token log-probabilities over yes/no or A/B/C/D tokens.

\paragraph{Action temporal localization.}
Industry temporal-localization questions use the official SFT checkpoint in two complementary branches.
The direct branch answers the original multiple-choice question with a standard industrial video MCQ prompt.
The temporal branch first selects frames around the candidate option timestamps and then applies the same deterministic MCQ decoding to this time-focused visual evidence.
When the two branches disagree, the final router treats option B as a low-confidence fallback: if exactly one branch predicts B, it returns the other branch; otherwise it returns the direct-branch answer.

\paragraph{Dominant held-object identification.}
Dominant held-object questions use the base checkpoint and an option-verification prompt for the object manipulated by the operator hand named in the question.
For each candidate object, the model is asked whether that object is the one the target hand predominantly interacts with across the provided segment.
If the effective frame sequence is longer than one model call, the frames are chunked, and each candidate is scored by a frame-weighted mean yes/no margin.
The option with the largest predominant-interaction score is selected.

\paragraph{Next interaction prediction.}
Next-interaction questions are answered with a pairwise tail-frame expert on the base checkpoint.
The router keeps the final frames of the observed clip, compares every pair of candidate future interactions using an A/B prompt, and runs the comparison symmetrically with the option order swapped.
For each pair, the two A/B log-probability margins are averaged into a margin for one option over the other.
The final score for each of the four choices is the sum of its pairwise margins, and the highest-scoring option is returned.

\paragraph{Object counting.}
Object-counting questions use three base-model experts.
Two experts ask the model to output one permille-coordinate bounding box for every visible label from a fixed industrial vocabulary: battery, battery connector, board, button, electric screwdriver, oscilloscope, oscilloscope component, pliers, power supply, power supply cables, screen, and screwdriver.
The first uses the global labeled-box prompt, while the second uses a plainer variant that asks for every clearly visible candidate label.
Both outputs are parsed into unique visible labels and mapped to the numeric answer option.
A third option-guided expert scores the four count options with yes/no verification.
The final router returns the labeled-box answer when the two box prompts agree; otherwise it falls back to the option-guided answer, except for one dense-count correction where both box prompts predict B, the option-guided branch predicts D, both box parsers see at least 11 unique labels, and D corresponds to a count of 8.

\paragraph{Object not visible identification.}
Object-not-visible questions use the official SFT checkpoint.
The prompt asks which of the four refined object categories is not visible in the chronologically ordered images and includes a special clarification that a button refers to a button on the electrical box or equipment panel when a button option appears.
Instead of generating a free-form answer, the router computes next-token log-probabilities for A, B, C, and D after the MCQ prompt and selects the letter with the highest probability.

\paragraph{Object spatial localization.}
Object spatial localization uses the base checkpoint with a coordinate-to-region strategy.
The router first restricts the evidence to the frame at the question timestamp by using a zero-radius timepoint neighborhood.
The model is then prompted to return exactly six approximate points on the referenced held object in permille image coordinates.
The six points are parsed, their median $x$ and $y$ coordinates are used as the predicted object location, and this point is mapped to the answer option whose canonical region center is nearest in squared Euclidean distance.

\subsection{XSports Domain}
\label{subsec:xsports_domain}

\paragraph{Common setup.}
For XSports, the final router fills only ExtrameSportFPV rows and normalizes every prediction to A, B, C, or D.
It loads the official sft checkpoint for all XSports question types and uses uniform sampling of up to eight frames per question.
For next direction prediction, action sequence identification, and sport identification, the default branch is option-guided verification: each answer choice is turned into a yes/no visual-support question, the model scores the next-token yes and no probabilities, and the option with the largest yes probability is selected.
This same option-guided branch is also computed for special-action questions because it can override the pairwise expert in narrow cases.

\paragraph{Special action identification.}
Special-action questions are primarily handled by a pairwise action expert.
The expert compares every pair of candidate actions with a task-specific A/B prompt that defines the visual requirements for actions such as spin, vault, and climb, scores both option orders with A/B next-token log-probabilities, and aggregates the symmetric pairwise margins over all six option pairs.
The pairwise winner is used by default.
The router falls back to the option-guided answer when the pairwise winner is Spin, or when the pairwise winner is Vault with margin below 5 and the option-guided answer is one of Jump, Climb, Flip, or Fly.

\paragraph{XSports action temporal localization.}
XSports temporal localization uses a multi-expert router because the candidate times are close and fast actions can be confused with preparation or continuation.
For each question, the main SFT transition expert uses a 15-second duration assumption, stamps the sampled frames with estimated timestamps, and scores every candidate time $t$ from two local windows: frames nearest to $t-1.2$, $t-0.6$, and $t-0.2$ ask whether the action has already started before $t$, while frames nearest to $t+0.2$, $t+0.6$, and $t+1.2$ ask whether the action has started by or just after $t$.
The candidate score is $\log p(\mathrm{post\ started}) + \log(1-p(\mathrm{pre\ started}))$, and the highest-scoring candidate gives the transition answer.
In parallel, a legacy SFT option-guided expert scores the original four timestamp options with temporal-anchor instructions.
When these two experts disagree and the transition margin, legacy margin, time gap, rank, and action type satisfy the router conditions, an A/B pairwise arbiter compares the two candidate times using a small frame panel around both times plus start/end anchors.
The final ATL answer is chosen by a heuristic meta-router that considers the conservative and aggressive transition-versus-legacy routes together with three extra SFT/base checks: an SFT no-anchor option-guided run, an SFT 30-second transition run, a base-model 15-second transition run with product scoring, and a base-model 30-second transition run.

\subsection{Final accuracy}
\label{subsec:final_accuracy}

Table~\ref{tab:final_accuracy} reports the final accuracy of our submitted system on each target domain and overall.

\begin{table}[t]
  \centering
  \small
  \setlength{\tabcolsep}{3.5pt}
  \caption{Final accuracy on EgoCross target domains.}
  \label{tab:final_accuracy}
  \begin{tabular}{@{}lccccc@{}}
    \toprule
    & Animal & XSports & Industry & Surgery & Overall \\
    \midrule
    Accuracy & 0.7705 & 0.6341 & 0.6449 & 0.6572 & 0.669801 \\
    \bottomrule
  \end{tabular}
\end{table}

\section{Conclusion}
\label{sec:conclusion}

We present a nearly training-free domain-wise inference strategy for the EgoCross source-limited challenge.
Instead of relying on a larger base model or extensive task-specific fine-tuning, our system keeps Qwen3-VL-4B~\cite{bai2025qwen3} fixed for surgery and animal domains, and uses only the official two-epoch SFT checkpoint for XSports and industry.
The central idea is to adapt the inference interface rather than the model parameters: each domain receives a task-aware procedure that exposes the relevant visual, temporal, and option-level cues to the VLM.
The final submission achieves 66.98\% overall accuracy, with 77.05\% on Animal, 63.41\% on XSports, 64.49\% on Industry, and 65.72\% on Surgery.
These results support the view that many errors in rare egocentric scenarios come from a transfer-interface mismatch rather than a complete absence of relevant knowledge in the base model.
A natural next step is to replace the hand-designed routing rules with a learned or self-adaptive controller while preserving the same principle of reducing unnecessary burden on the VLM.
{
    \small
    \bibliographystyle{ieeenat_fullname}
    \bibliography{main}
}


\end{document}